%% file: main.tex
\documentclass[10pt,twocolumn,letterpaper]{article}
\usepackage{cvpr}

\definecolor{cvprblue}{rgb}{0.21,0.49,0.74}
\usepackage[pagebackref,breaklinks,colorlinks,citecolor=cvprblue]{hyperref}
\usepackage{pifont}
\usepackage{colortbl}
\usepackage{multirow}

\newcommand{\method}{PartGen\xspace}

\def\paperID{5026}
\def\confName{CVPR}
\def\confYear{2025}

\makeatletter
\renewcommand{\paragraph}{%
    \@startsection{paragraph}{4}%
    {\z@}{-0.5em}{-0.5em}%
    {\normalfont\normalsize\bfseries}%
}
\makeatother

\title{\method: Part-level 3D Generation and Reconstruction \\ with Multi-View Diffusion Models}

\author{Minghao Chen$^{1,2}$ \quad Roman Shapovalov$^{2}$ \quad Iro Laina$^{1}$ \quad Tom Monnier$^{2}$  \\ Jianyuan Wang$^{1,2}$ \quad David Novotny$^{2}$ \quad Andrea Vedaldi$^{1,2}$ \\
$^1$Visual Geometry Group, University of Oxford \quad $^2$Meta AI \\
\href{https://silent-chen.github.io/PartGen/}{\tt\small {\nolinkurl{silent-chen.github.io/PartGen}}}
}
\begin{document}

\twocolumn[{
\maketitle
\begin{center}
\vspace{-1.6em}
\includegraphics[width=\linewidth]{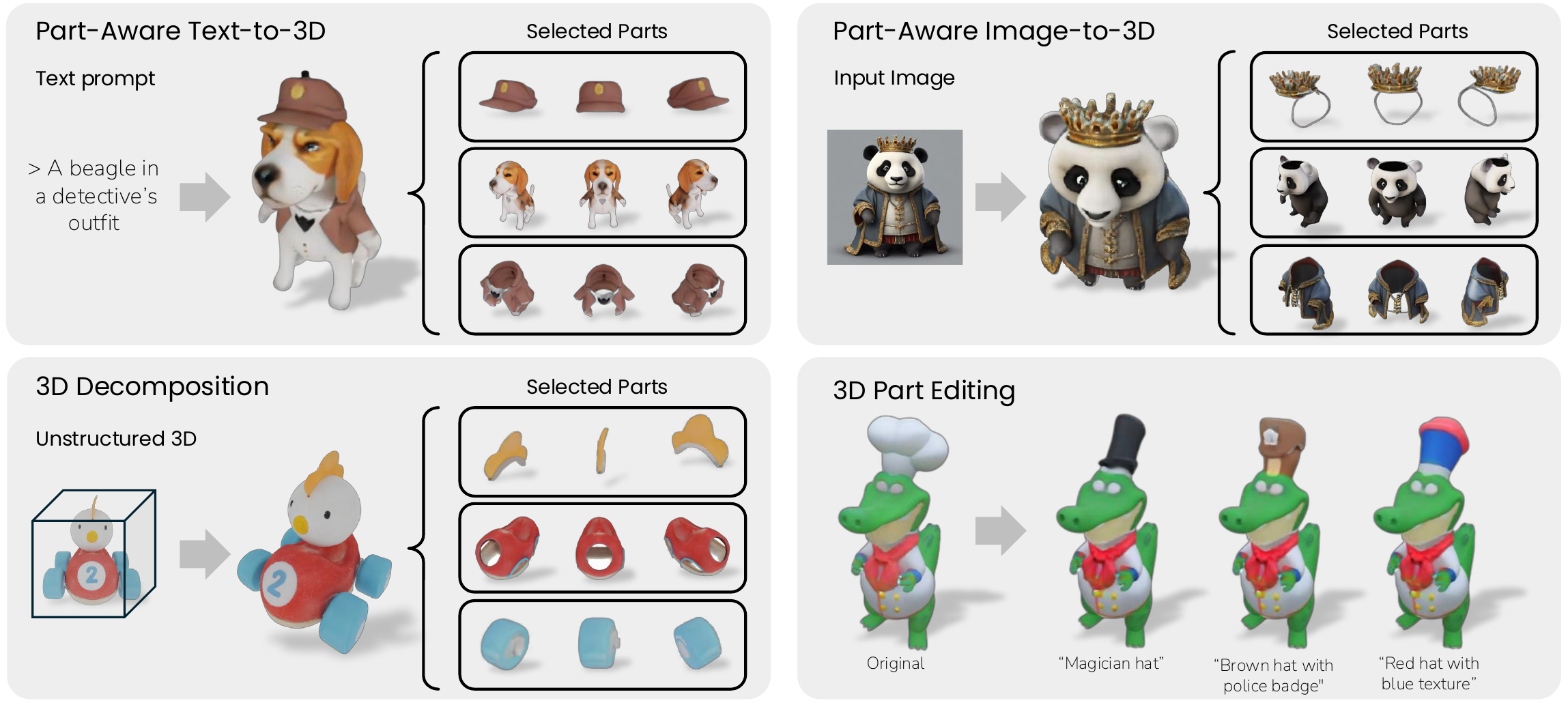}
\end{center}
\vspace{-1.2em}
\captionsetup{type=figure}
\captionof{figure}{
We introduce \textbf{\method}, a pipeline that generates compositional 3D objects similar to a human artist.
It can start from text, an image, or an existing, unstructured 3D object.
It consists of a multi-view diffusion model that identifies plausible parts automatically and another that completes and reconstructs them in 3D, accounting for their context, \ie, the other parts, to ensure that they fit together correctly. Additionally, \method enables 3D part editing based on text instructions, enhancing flexibility and control in 3D object creation.
}\label{fig:teaser}
\vspace{1.35em}
}]

\newcommand\blfootnote[1]{%
\begingroup 
\renewcommand\thefootnote{}\footnote{#1}%
\addtocounter{footnote}{-1}%
\endgroup 
}
{
    \blfootnote{
    Work completed during Minghao’s internship at Meta.
}
}

\input{sec/0_abstract}
\input{sec/1_intro}

\input{sec/2_relwork}

\input{sec/3_method}
\input{sec/4_experiments}
\input{sec/5_conclusion}

{
    \small
    \bibliographystyle{ieeenat_fullname}
    \bibliography{main,vedaldi_general,vedaldi_specific}
}
\input{sec/X_suppl}

\end{document}

%% file: sec/0_abstract.tex
\begin{abstract}
Text- or image-to-3D generators and 3D scanners can now produce 3D assets with high-quality shapes and textures.
These assets typically consist of a single, fused representation, like an implicit neural field, a Gaussian mixture, or a mesh, without any useful structure.
However, most applications and creative workflows require assets to be made of several meaningful parts that can be manipulated independently.
To address this gap, we introduce \method, a novel approach that generates 3D objects composed of meaningful parts starting from text, an image, or an unstructured 3D object.
First, given multiple views of a 3D object, generated or rendered, a multi-view diffusion model extracts a set of plausible and view-consistent part segmentations, dividing the object into parts.
Then, a second multi-view diffusion model takes each part separately, fills in the occlusions, and uses those completed views for 3D reconstruction by feeding them to a 3D reconstruction network.
This completion process considers the context of the entire object to ensure that the parts integrate cohesively.
The generative completion model can make up for the information missing due to occlusions; in extreme cases, it can hallucinate entirely invisible parts based on the input 3D asset.
We evaluate our method on generated and real 3D assets and show that it outperforms segmentation and part-extraction baselines by a large margin.
We also showcase downstream applications such as 3D part editing.
\end{abstract}

%% file: sec/1_intro.tex
\section{Introduction}%
\label{sec:intro}

High-quality textured 3D assets can now be obtained through generation from text or images~\cite{tripo3d24text-to-3d,siddiqui24meta,meshy24text-to-3d,lumaai24genie,melas-kyriazi24im-3d,gao24cat3d:,deemos24rodin,csm24csm-text-to-3d}, or through photogrammetry techniques~\cite{downs2022google, wu2023omniobject3d, pan2023aria}.
However, the resulting objects are \emph{unstructured}, consisting of a single, monolithic representation, such as an implicit neural field, a mixture of Gaussians, or a mesh.
This is not good enough in a professional setting, where the \emph{structure} of an asset is also of paramount importance.
While there are many aspects to the structure of a 3D object (\eg, the mesh topology), parts are especially important as they enable reuse, editing and animation.

In this paper, we thus consider the problem of obtaining \emph{structured} 3D objects that are formed by a collection of meaningful \emph{parts}, akin to the models produced by human artists.
For example, a model of a person may be decomposed into its clothes and accessories, as well as various anatomical features like hair, eyes, teeth, limbs, etc.
However, if the object is generated or scanned, different parts are usually `fused' together, missing the internal surfaces and the part boundaries.
This means that physically detachable parts appear glued together, with a jarring effect.
Furthermore, parts carry important information and functionality that those models lack.
For example, different parts may have distinct animations or different materials.
Parts can also be replaced, removed, or edited independently.
For instance, in video games, parts are often reconfigured dynamically, \eg, to represent a character picking up a weapon or changing clothes.
Due to their semantic meaning, parts are also important for 3D understanding and applications like robotics, embodied AI, and \emph{spatial intelligence} \cite{cpm2023, mees23hulc2}. 

Inspired by these requirements, we introduce \method, a method to upgrade existing 3D generation pipelines from producing unstructured 3D objects to generating objects as compositions of meaningful 3D parts.
To do this, we address two key questions:
(1) how to automatically \emph{segment} a 3D object into parts, and (2) how to extract high-quality, \emph{complete} 3D parts even when these are only partially—or not at all—visible from the exterior of the 3D object.

Crucially, both part segmentation and completion are highly ambiguous tasks. 
First, since different artists may find it useful to decompose the same object in different ways, there is no `gold-standard' segmentation for any given 3D object.
Hence, a segmentation method should model the distribution of plausible part segmentations rather than a single one.
Second, current 3D reconstruction and generation methods only model an object's visible outer surface, omitting inner or occluded parts.
Therefore, decomposing an object into parts often requires completing these parts or even entirely hallucinating them.

To model this ambiguity, we base part segmentation and reconstruction on 3D generative models.
We note that most state-of-the-art 3D generation pipelines~\cite{li24instant3d:,tripo3d24text-to-3d,siddiqui24meta,meshy24text-to-3d,lumaai24genie,melas-kyriazi24im-3d,gao24cat3d:,deemos24rodin,csm24csm-text-to-3d} start by generating several consistent 2D views of the object, and then apply a 3D reconstruction network to those images to recover the 3D object.
We build upon this two-stage scheme to address both part segmentation and reconstruction ambiguities.

In the first stage, we cast part segmentation as a \emph{stochastic multi-view-consistent colouring problem}, leveraging a multi-view image generator fine-tuned to produce colour-coded segmentation maps across multiple views of a 3D object.
We do not assume any explicit or even deterministic taxonomy of parts; the segmentation model is learned from a large collection of artist-created data, capturing how 3D artists decompose objects into parts.
The benefits of this approach are twofold. 
First, it leverages an image generator which is already trained to be view-consistent.
Second, a generative approach allows for multiple plausible segmentations by simply re-sampling from the model. 
We show that this process results in better segmentation than that obtained by fine-tuning a model like SAM~\cite{kirillov23segment} or SAM2~\cite{ravi24sam-2} for the task of multi-view segmentation:
while the latter can still be used, our approach better captures the artists' intent.

For the second problem, namely reconstructing a segmented part in 3D, an obvious approach is to mask the part within the available object views, and then use a 3D reconstructor network to recover the part in 3D.
However, when the part is heavily occluded, this task amounts to \emph{amodal reconstruction}, which
is highly ambiguous and thus badly addressed by the deterministic reconstructor network.
Instead, and this is our core contribution, we propose to tune another multi-view generator to \emph{complete} the views of the part while \emph{accounting for the context} of the object as a whole.
In this manner, the parts can be reconstructed reliably even if they are only partially visible, or even not visible, in the original input views.
Furthermore, the resulting parts fit together well and, when combined, form a coherent 3D object.

We show that \method can be applied to different input modalities.
Starting from text, an image, or a areal-world 3D scan, \method can generate 3D assets with meaningful parts.
We assess our method empirically on a large collection of 3D assets produced by 3D artists or scanned, both quantitatively and qualitatively.
We also demonstrate that \method can be easily extended to the 3D part editing task.

%% file: sec/2_relwork.tex
\section{Related Work}%
\label{sec:relwork}

\paragraph{3D generation from text and images.}

The problem of generating 3D assets from text or images has been thoroughly studied in the literature.
Some authors have built generators from scratch.
For instance, CodeNeRF~\cite{jang21codenerf:} learns a latent code for NeRF in a Variational Autoencoder fashion, and Shap-E~\cite{jun23shap-e:} and 3DGen~\cite{gupta233dgen:} does so using latent diffusion, $\textrm{PC}^2$~\cite{melas-kyriazi23pc2} and Point-E~\cite{nichol22point-e:} diffuse a point cloud, and MosaicSDF a semi-explicit SDF-based representation~\cite{yariv23mosaic-sdf}.
However, 3D training data is scarce, which makes it difficult to train text-based generators directly.

DreamFusion~\cite{poole23dreamfusion:} demonstrated for the first time that 3D assets can be extracted from T2I diffusion models with \emph{Score Distillation Sampling} (SDS) loss.
Variants of DreamFusion explore representations like hash grids~\cite{lin22magic3d:, qian23magic123:}, meshes~\cite{lin22magic3d:} and 3D Gaussians (3DGS)~\cite{tang23dreamgaussian:, yi23gaussiandreamer:, chen23text-to-3d}, tweaks to the SDS loss~\cite{wang23score, wang23prolificdreamer:, zhu23hifa:, huang23dreamtime:}, conditioning on an input image~\cite{melas-kyriazi23realfusion, qian23magic123:, tang23make-it-3d:, yu23hifi-123:, sun23dreamcraft3d:}, and regularizing normals or depth~\cite{qiu23richdreamer:,sun23dreamcraft3d:, shi24mvdream:}. 

Other works focus on improving the 3D awareness of the T2I model, simplifying extracting a 3D output and eschewing the need for slow SDS optimization.
Inspired by 3DIM~\cite{watson23novel}, Zero-1-to-3~\cite{liu23zero-1-to-3:} fine-tunes the 2D generator to output novel views of the object.
Two-stage approaches~\cite{liu23one-2-3-45:, long23wonder3d:, liu23syncdreamer:, yang23dreamcomposer:, yang23consistnet:, han2025vfusion3d, chan23generative, tang24mvdiffusion:, hollein24viewdiff:, melas-kyriazi24im-3d, chen24v3d:,gao24cat3d:, xu24dmv3d:, han2024flex3d, shi24mvdream:, wang24imagedream:,xu24instantmesh:} take the output of a text- or image-to-multi-view model that generates multiple views of the object and reconstruct the latter using multi-view reconstruction methods like NeRF~\cite{mildenhall20nerf:} or 3DGS~\cite{kerbl233d-gaussian}.
Other approaches reduce the number of input views generated and learn a fast feed-forward network for 3D reconstruction.
Perhaps the most notable example is Instant3D~\cite{li24instant3d:} based on the \emph{Large Reconstruction Model} (LRM)~\cite{hong24lrm:}. Recently, there are works focusing on 3D compositional generation \cite{cohen-bar23set-the-scene:, Po2023Compositional3S, yan2024dreamdissector, li2023focaldreamer}. D3LL~\cite{epstein2024disentangled3dscenegeneration} learns 3D object composition through distilling from a 2D T2I generator. ComboVerse~\cite{chen2024comboverse} starts from a single image, but mostly at the levels of different objects instead of their parts, performs single-view inpainting and reconstruction, and uses SDS optimization for composition.

\paragraph{3D segmentation.}

Our work decomposes a given 3D object into parts.
Several works have considered segmenting 3D objects or scenes represented in an unstructured manner, lately as neural fields or 3D Gaussian mixtures.
Semantic-NeRF~\cite{zhi21in-place} was the first to fuse 2D semantic segmentation maps in 3D with neural fields.
DFF~\cite{kobayashi22decomposing} and
N3F~\cite{tschernezki22neural}
propose to map 2D features to 3D fields, allowing their supervised and unsupervised segmentation.
LERF~\cite{kerr23lerf:} extends this concept to language-aware features like CLIP~\cite{radford21learning}.
Contrastive Lift~\cite{bhalgat23contrastive} considers instead instance segmentation, fusing information from several independently-segmented 
views using a contrastive formulation.
GARField~\cite{kim24garfield:} and OminiSeg3D~\cite{ying2024omniseg3d} consider that concepts exist at different levels of scale, which they identify with the help of SAM~\cite{kirillov23segment}.
LangSplat~\cite{qin24langsplat:} leverages both CLIP and SAM, creating distinct 3D language fields to model each SAM scale explicitly, while N2F2~\cite{bhalgat24n2f2} automates binding the correct scale to each concept.
Neural Part Priors~\cite{bokhovkin2023neural} completes and decomposes 3D scans with learned part priors in a test-time optimization manner. 
Finally, Uni3D~\cite{zhou24uni3d:} learns a `foundation' model for 3D point clouds 
that can perform zero-shot segmentation. 

\paragraph{Primitive-based representations.}

Some authors proposed to represent 3D objects as a mixture of primitives~\cite{zhan2020generative}, which can be seen as related to parts, although they are usually non-semantic.
For example, SIF~\cite{genova19learning} represents an occupancy function as a 3D Gaussians mixture.
LDIF~\cite{genova20local} uses the Gaussians to window local occupancy functions implemented as neural fields~\cite{mescheder19occupancy}.
Neural Template~\cite{hui22neural} and SPAGHETTI~\cite{amir22spaghetti:} learn to decompose shapes in a similar manner using an auto-decoding setup.
SALAD~\cite{koo23salad:} uses SPAGHETTI as the latent representation for a diffusion-based generator.
PartNeRF~\cite{tertikas23partnerf:} is conceptually similar, but builds a mixture of NeRFs.
NeuForm~\cite{lin22neuform:} and DiffFacto~\cite{nakayama23difffacto:}
learn representations that afford part-based control. DBW~\cite{monnier2023differentiable} decomposes real-world scenes with textured superquadric primitives. 

\paragraph{Semantic part-based representations.}

Other authors have considered 3D parts that are semantic.
PartSLIP~\cite{liu23partslip:} and PartSLIP++~\cite{zhou23partslip:} use vision-language model to segment objects into parts using point clouds as representation.
Part123~\cite{liu24part123:} is conceptually similar to Contrastive Lift~\cite{bhalgat23contrastive}, but applied to object than scenes, and to the output of a monocular reconstruction network instead of NeRF\@.

In this paper, we address a problem different from the ones above.
We generate compositional 3D objects from various modalities using multi-view diffusion models for segmentation and completion.
Parts are meaningfully segmented, fully reconstructed, and correctly assembled.
We handle the ambiguity of these tasks in a generative way.

%% file: sec/3_method.tex
\section{Method}%
\label{sec:method}

\newcommand{\object}{\mathbf{L}}
\newcommand{\objectpart}{\mathbf{S}}

\input{figures/fig_overview}

This section introduces \method, our framework for generating 3D objects that are fully decomposable into \emph{complete} 3D parts.
Each part is a distinct, human-interpretable, and self-contained element, representing the 3D object compositionally.
\method can take different modalities as input (text prompts, image prompts, or 3D assets) and performs part segmentation and completion by repurposing a powerful multi-view diffusion model for these two tasks.
An overview of \method is shown in \Cref{fig:overview}.

The rest of the section is organised as follows.
In \cref{sec:prelmimaries}, we introduce the necessary background on multi-view diffusion and how \method can be applied to text, image, or 3D model inputs briefly.
Then, in \cref{sec:part-segmentation,sec:part-completion,sec:part-reconstruction} we describe how \method automatically segments, completes, and reconstructs meaningful parts in 3D.

\subsection{Background on 3D generation}%
\label{sec:prelmimaries}

First, we provide essential background on multi-view diffusion models for 3D generation~\cite{shi24mvdream:,li24instant3d:,siddiqui24meta}.
These methods usually adopt a two-stage approach to 3D generation.

In the first stage, given a prompt $y$, an image generator $\Phi$ outputs several 2D views of the object from different vantage points.
Depending on the nature of $y$, the network $\Phi$ is either a text-to-image (T2I) model~\cite{shi24mvdream:, li24instant3d:} or a image-to-image (I2I) one~\cite{wang24imagedream:, shi23zero123:}.
These are fine-tuned to output a single `multi-view' image $I \in \mathbb{R}^{3\times 2H \times 2W}$, where views from the four cardinal directions around the object are arranged into a $2 \times 2$ grid.
This model thus provides a probabilistic mapping $I \sim p(I \mid \Phi,y)$.
The 2D views $I$ are subsequently passed to a Reconstruction Model (RM)~\cite{li24instant3d:,siddiqui24meta,xu24grm:} $\Psi$, \ie, a neural network that reconstructs the 3D object $\object$ in both shape and appearance.
Compared to direct 3D generation, this two-stage paradigm takes full advantage of an image generation model pre-trained on internet-scale 2D data.

This approach is general and can be applied with various implementations of image-generation and reconstruction models.
Our work in particular follows a setup similar to AssetGen~\cite{siddiqui24meta}.
Specifically, we obtain $\Phi$ by finetuning a pre-trained text-to-image diffusion model with an architecture similar to Emu~\cite{dai23emu:}, a diffusion model in a 8-channel latent space, the mapping to which is provided by a specially trained variational autoencoder (VAE).
The detailed fine-tuning strategy can be found in \cref{sec:applications} and supplementary material.
When the input is a 3D model, we render multiple views to form the grid view.
For the RM $\Psi$ we use LightplaneLRM~\cite{cao2024lightplane}, trained on our dataset.

\subsection{Multi-view part segmentation}%
\label{sec:part-segmentation}

The first major contribution of our paper is a method for segmenting an object into its constituent parts.
Inspired by multi-view diffusion approaches, we frame object decomposition into parts as a \emph{multi-view segmentation} task, rather than as direct 3D segmentation.
At a high-level, the goal is to map $I$ to a collection 2D masks $M^1,\dots,M^S \in \{0,1\}^{2H\times 2W}$, one for each visible part of the object.
Both image $I$ and masks $M_i$ are multi-view grids.

Addressing 3D object segmentation through the lens of multi-view diffusion offers several advantages.
First, it allows us to repurpose existing multi-view models $\Phi$, which, as described in \cref{sec:prelmimaries}, are already pre-trained to produce multi-view consistent generations in the RGB domain.
Second, it integrates easily with established multi-view frameworks.
Third, decomposing an object into parts is an inherently non-deterministic, ambiguous task as it depends on the desired verbosity level, individual preferences, and artistic intent.
By learning this task with probabilistic diffusion models, we can effectively capture and model this ambiguity.
We thus train our model on a curated dataset of artist-created 3D objects, where each object $\object$ is annotated with a possible decomposition into 3D parts, $\object = (\objectpart^1, \dots, \objectpart^S)$. The dataset details are provided in \cref{sec:trainign-data}.

Consider that the input is a multi-view image $I$, and the output is a set of multi-view part masks $M^1,M^2,\dots, M^S$.
To finetune our multi-view image generators $\Phi$ for mask prediction, we quantize the RGB space into $Q$ different colors $c_1,\dots,c_Q \in [0,1]^3$.
For each training sample $\object = (\objectpart^k)_{k=1}^S$, we assign colors to the parts, mapping part $\objectpart^k$ to color $c_{\pi_k}$, where $\pi$ is a random permutation on $\{1,\dots,Q\}$ (we assume that $Q \geq S$).
Given this mapping, we render the segmentation map as a multi-view RGB image $C \in [0,1]^{3\times 2H\times 2W}$ (\cref{fig:visualization_segmentation}).
Then, we fine-tune $\Phi$ to
(1) take as conditioning the multi-view image $I$, and
(2) to generate the color-coded multi-view segmentation map $C$, hence sampling a distribution $C \sim p(C \mid \Phi_\text{seg}, I)$.

This approach can produce alternative segmentations by simply re-running $\Phi_\text{seg}$, which is stochastic.
It further exploits the fact that $\Phi_\text{seg}$ is stochastic to discount the specific `naming' or coloring of the parts, which is arbitrary.
Naming is a technical issue in instance segmentation which usually requires ad-hoc solutions, and here is solved `for free'.

To extract the segments at test time, we sample the image $C$ and simply quantize it based on the reference colors $c_1,\dots,c_Q$, discarding parts that contain only a few pixels.

\input{figures/fig_data_visualization}

\paragraph{Implementation details.}

The network $\Phi_\text{seg}$ has the same architecture as the network $\Phi$ with some changes to allow conditioning on the multi-view image $I$: we encode it into latent space with the VAE and stack it with the noised latent as the input to the diffusion network.

\subsection{Contextual part completion}%
\label{sec:part-completion}

The method so far has produced a multi-view image $I$ of the 3D object along with 2D segments $M^1,M^2,\dots, M^S$.
What remains is to convert those into the \emph{full} 3D part reconstructions.
Given a mask $M$, in principle we could simply submit the masked image $I \odot M$ to the RM $\Psi$ to obtain a 3D reconstruction of the part, \ie, $\hat \objectpart = \Psi(I \odot M)$.
However, in multi-view images, some parts can be heavily occluded by other parts and, in extreme cases, entirely invisible.
While we could train the RM to handle such occlusions directly, in practice this does not work as part completion is inherently a stochastic problem, whereas the RM is deterministic.

To handle this ambiguity, we repurpose yet again the multi-view generator $\Phi$, this time to perform part completion.
The latter model is able to generate a 3D object from text or single image, so, properly fine-tuned, it should be able to hallucinate any missing portion of a part.

Formally, we consider fine-tuning $\Phi$ to sample a view $J \sim p(J \mid I\odot M)$, mapping the masked image $I\odot M$ to the completed multi-view image $J$ of the part.
However, we note that sometimes parts are barely visible, so the masked image $I\odot M$ provides very little information.
Furthermore, we need the generated part to \emph{fit well with the other parts and the whole object}.
Hence, we provide to the model also the un-masked image $I$ for \emph{context}.
Thus, condition  $p(J \mid I\odot M, I, M)$ on the masked image $I\odot M$, the unmasked image $I$, and the mask $M$.
The importance of the context $I$ increases with the extent of the occlusion.

\paragraph{Implementation details.}

The network architecture resembles that of \cref{sec:part-segmentation}, but extends the conditioning, motivated by the inpainting setup in \cite{rombach22high-resolution}.
We apply the pre-trained VAE separately to the masked image $I\odot M$ and context image $I$, yielding $2 \times 8$ channels, and stack them with the 8D noise image and the unencoded part mask $M$ to obtain the 25-channel input to the diffusion model.
Example results are shown in \Cref{fig:part_completion}.

\subsection{Part reconstruction}%
\label{sec:part-reconstruction}

Given a multi-view part image $J$, the final step is to reconstruct the part in 3D.
Because the part views are now complete and consistent, we can simply use the RM to obtain a predicted reconstruction $\hat \objectpart = \Psi(J)$ of the part.
We found that the model does not require special finetuning to move from objects to their parts, so any good quality reconstruction model can be plugged into our pipeline directly.

\subsection{Training data}%
\label{sec:trainign-data}

To train our models, we require a dataset of 3D models consisting of multiple parts.
We have built this dataset from a collection of 140k 3D-artist generated assets that we licensed for AI training from a commercial source.
Each asset $\object$ is stored as a GLTF scene that contains, in general, several watertight meshes $(\objectpart^1,\dots,\objectpart^S)$ that often align with semantic parts due to being created by a human who likely aimed to create an editable asset. Example objects from the dataset are shown in \cref{fig:data_visualization}. We preprocess data differently for each of the three models we fine tuned.

\paragraph{Multi-view generator data.}
To train the multi-view generator models $\Phi$, first of all, we have to render the target multi-view images $I$ consisting of 4 views to the full object.
Following Instant3D\,\cite{li24instant3d:}, we rendered shaded colours $I$ from the 4 views from the orthogonal azimuths and 20$^\circ$ elevation and arranged them in a $2 \times 2$ grid.
In case of \emph{text conditioning}, training data consists of the pairs $\{(I_n, y_n)\}_{n=1}^N$ of multi-view images and their text captions
Following AssetGen~\cite{siddiqui24meta}, we choose 10k highest quality assets and generated their text captions using CAP3D-like pipeline~\cite{luo2023scalable} that used LLAMA3 model~\cite{dubey24the-llama}.
In case of \emph{image conditioning}, we use all 140k models, and the conditioning $y_n$ comes in form of single renders from a randomly sampled direction (not just one of the four used in $I_n$).

\input{figures/fig_visualization_segmentation}

\paragraph{Part segmentation and completion data.}
To train the part segmentation and completion networks, we need to additionally render the multi-view part images and their depth maps.
Since different creators have different ideas on part decomposition, we filter the dataset to avoid having excessively granular parts which likely lack semantic meaning.
To this end, we first cull the parts that take less than 5\% of the object volume, and then remove the assets that have more than 10 parts or consist of a single monolithic part.
This results in the dataset of 45k objects contain the total of 210k parts.
Given the asset $\object = (\objectpart^1,\dots,\objectpart^S)$, we render a set of multi-view images $\{J ^s\}_{s=1}^S$ (shown in \cref{fig:data_visualization}) and the corresponding depth maps $\{\delta ^s\}_{s=1}^S$ from the same viewpoints as above.

The \emph{segmentation diffusion network} is trained on the dataset of pairs $\{(I_n, \mathbf{M}_n)\}_{n=1}^N$, where the segmentation map $\mathbf{M} = [M^k]_{k=1}^S$ is a stack of multi-view binary part masks $M^k \in \{0, 1\}^{2H \times 2W}$.
Each mask shows the pixels where the appropriate part is visible in $I$:
$M^k_{i,j} = [k = \textrm{argmin}_l \delta^l_{i,j}]$, where $k, l \in \{1, \dots, S\}$ and brackets denote Iverson brackets.
The \emph{part completion network} is trained on the dataset of triplets $\{(I_{n'}, J_{n'}, M_{n'})\}_{{n'}=1}^{N'}$.
All the components are produces in the way described above.

%% file: figures/fig_overview.tex
\begin{figure*}[t]
\centering
\includegraphics[width=0.97\linewidth]{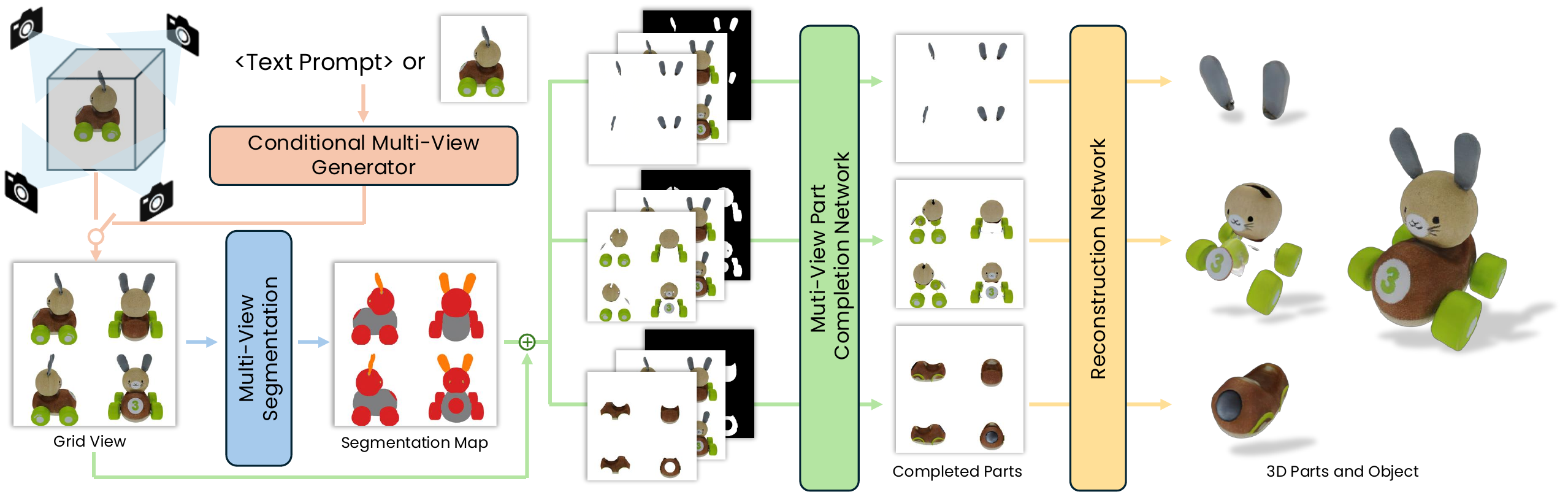}
\vspace{-.5em}
\caption{\textbf{Overview of \method.} Our method begins with text, single images, or existing 3D objects to obtain an initial grid view of the object. This view is then processed by a diffusion-based segmentation network to achieve multi-view consistent part segmentation. Next, the segmented parts, along with contextual information, are input into a multi-view part completion network to generate a fully completed view of each part. Finally, a pre-trained reconstruction model generates the 3D parts.}%
\label{fig:overview}
\vspace{-4mm}
\end{figure*}

%% file: figures/fig_data_visualization.tex
\begin{figure}[t]
\centering
\includegraphics[width=0.9\linewidth]{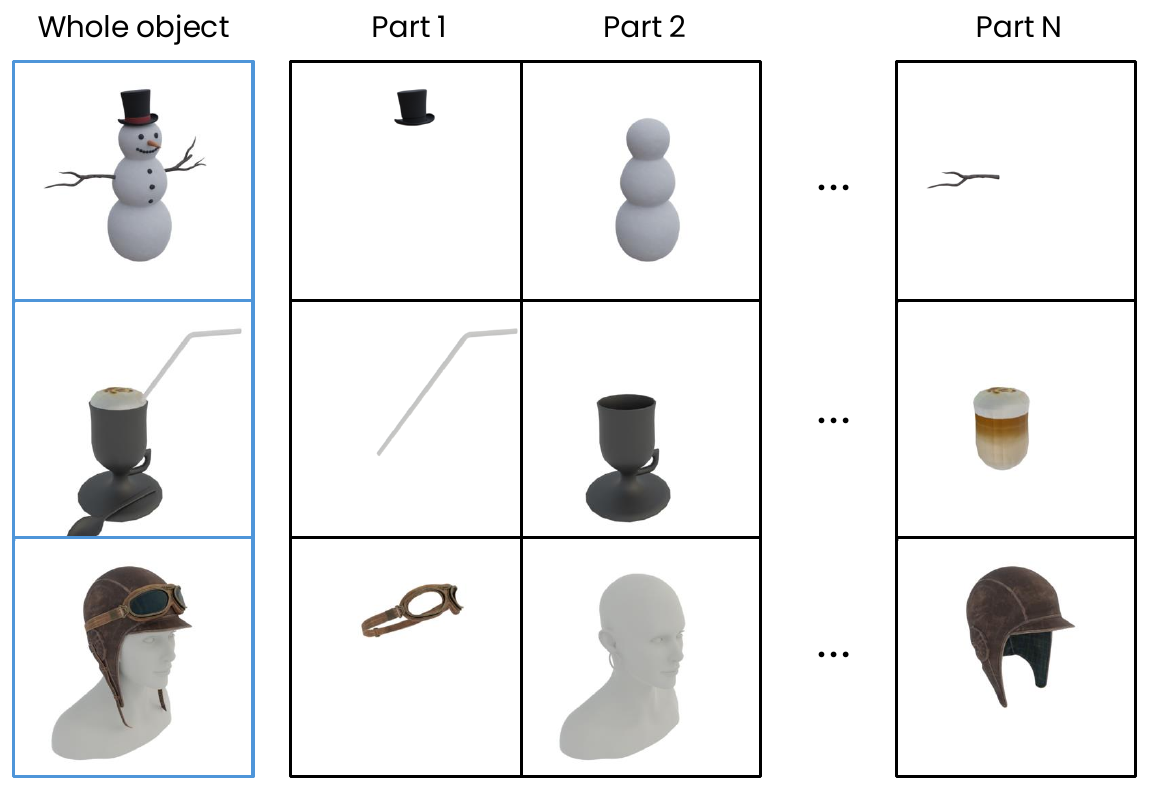}\vspace{-.3em}
\caption{\textbf{Training data.} 
We obtain a dataset of 3D objects decomposed into parts from assets created by artists.
These come `naturally' decomposed into parts according to the artist's design.
}%
\label{fig:data_visualization}
\vspace{-3mm}
\end{figure}

%% file: figures/fig_visualization_segmentation.tex
\begin{figure*}[t]
\centering
\includegraphics[width=0.95\linewidth]{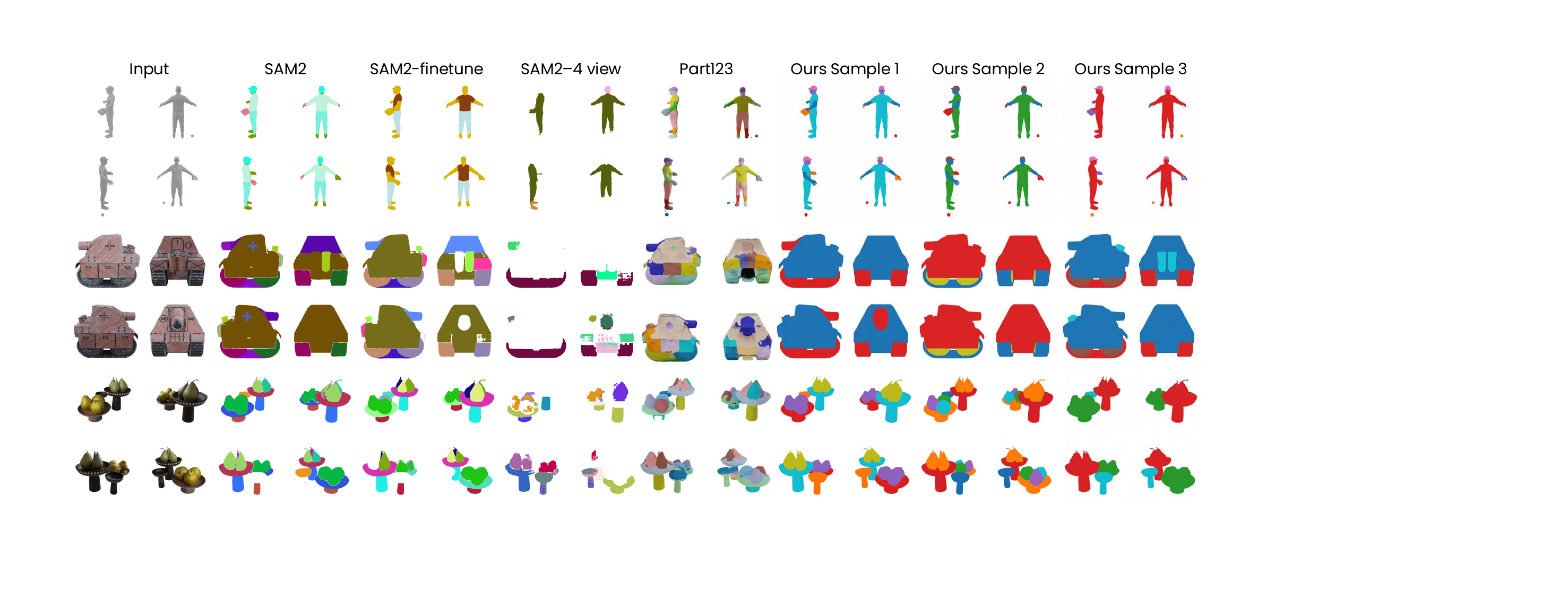}
\vspace{-2mm}

\caption{\textbf{Examples of automatic multi-view part segmentations.}
By running our method several times, we obtain different segmentations, covering the space of artist intents.}%
\label{fig:visualization_segmentation}
\vspace{-3mm}
\end{figure*}

%% file: sec/4_experiments.tex
\section{Experiments}%
\label{sec:experiments}

\input{tables/tab_eval_part_segmentation}

\input{tables/tab_eval_part_completion}

\input{figures/fig_part_completion}
\input{figures/fig_application}
\input{figures/fig_3d_editing}
\paragraph{Evaluation protocol.}

We first individually evaluate the two main components of our pipeline, namely part segmentation (\cref{sec:eval-part-segmentation}) and part completion and reconstruction (\cref{sec:eval-part-completion}).
We then evaluate how well the decomposed reconstruction matches the original object (\cref{sec:eval-part-composition}).
For all experiments, we use the held out 100 objects from the dataset described in \cref{sec:trainign-data}.

\subsection{Part segmentation}%
\label{sec:eval-part-segmentation}

\paragraph{Evaluation protocol.}

We set up two settings for the segmentation tasks. One is \emph{automatic part segmentation}, where the input is the multi-view image $I$ and requires the method to output all parts of the object $ M^{1},\ldots,M^{S}$. The other is \emph{seeded segmentation}, where we assume that users give a point as an additional input for a specific mask.
Now the segmentation algorithm is regarded as a black box
$
\mathbf{\hat M}=\mathcal{A}(I)
$
mapping the multi-view image $I$ to a ranked list of $N$ part segmentations (which can in general partially overlap).
This ranked list is obtained by scoring candidate regions and removing redundant ones. 
See the sup.~mat.~for more details.
We then match these segments to the ground-truth segments $M_k$ and report \emph{mean Average Precision} (mAP).
This precision can be low in practice due to the inherent ambiguity of the problem: many of the parts predicted by the algorithm will not match any particular artist's choice.

\paragraph{Baselines.}

We consider the original and fine-tuned SAM2~\cite{ravi24sam-2} as our baselines for multi-view segmentation.
We fine-tune SAM2 in two different ways. First, we fine-tune SAM2's mask decoder on our dataset, given the ground truth masks and randomly selected seed points for different views.
Second, we concatenate the four orthogonal views in a multi-view image $I$ and fine-tune SAM2 to predict the multi-view mask $\mathbf M$ (in this case, the seed point randomly falls in one of the views).
SAM2 produces three regions for each input image and seed point.
For automatic segmentation, we seed SAM2 with a set of query points spread over the object, obtaining three different regions for each seed point.
For seeded segmentation, we simply return the regions that SAM2 outputs for the given seed point. We also provide a comparison with recent work, Part123~\cite{liu24part123:}.
\paragraph{Results.}

We report the results in \cref{tab:eval-part-segmentation}.
As shown in the table, mAP results for our method are \emph{much} higher than others, including SAM2 fine-tuned on our data.
This is primarily because of the ambiguity of the segmentation task, which is better captured by our generator-based approach.
We further provide qualitative results in \cref{fig:visualization_segmentation}.

\subsection{Part completion and reconstruction}%
\label{sec:eval-part-completion}

We utilize the same test data as in \cref{sec:eval-part-segmentation}, forming tuples $(\objectpart,I,M^k,J^k)$ consisting of the 3D object part $\objectpart$, the full multi-view image $I$, the part mask $M^k$ and the multi-view image $J^k$ of the part, as described in \Cref{sec:trainign-data}. We choose one random part index $k$ per model, and will omit it from the notation below to be more concise.

\paragraph{Evaluation protocol.} The completion algorithm and its baselines are treated as a black box $\hat J = \mathcal{B}(I\odot M, I)$ that %
predicts the completed multi-view image $\hat J$.
We then compare $\hat J$ to the ground-truth render $J$ using
Peak Signal to Noise Ratio (PSNR) of the foreground pixels,
Learned Perceptual Image Patch Similarity (LPIPS)~\cite{zhang18the-unreasonable}, and CLIP similarity~\cite{radford21learning}. The latter is an important metric since the completion task is highly ambiguous, and thus evaluating \emph{semantic} similarity can provide additional insights. %
We also evaluate the quality of the reconstruction of the predicted completions by comparing the reconstructed object part $\hat\objectpart = \Phi(\hat J)$ to the ground-truth part $\objectpart$ using the same metrics, but averaged after rendering the part to four random novel viewpoints.

\paragraph{Results.}
We compare our part completion algorithm ($\hat J = \mathcal{B}(I\odot M, I)$) to several baselines and the oracle, testing
using no completion ($\hat J = I\odot M$),
omitting context ($\hat J = \mathcal{B}(I\odot M)$),
completing single views independently ($\hat J_v = \mathcal{B}(I_v\odot M_v, I_v)$),
and the oracle ($\hat J = J$).
The latter provides the upper-bound on the part reconstruction performance, where the only bottleneck is the RM.

As shown in the table \cref{tab:eval-part-completion}, our model largely surpasses the baselines.
Both joint multi-view reasoning and contextual part completion are important for good performance.
We further provide qualitative results in \cref{fig:part_completion}.

\subsection{Reassembling parts}%
\label{sec:eval-part-composition}
\input{tables/tab_eval_part_composition}

\paragraph{Evaluation protocol.}

Starting from multi-view image $I$ of a 3D object $\object$, we run the segmentation algorithm to obtain segmentation $(\hat M^1,\dots,\hat M^S)$, reconstruct each 3D part as
$\hat \objectpart^k = \Phi(\hat J^k)$, and reassemble the 3D object $\hat{\object}$ by merging the 3D parts $\{\hat{\objectpart}^1, \dots, \hat{\objectpart}^N\}$.
We then compare $\hat{\object} = \bigcup_k \Phi(\hat J_k)$  to the unsegmented reconstruction $\hat \object = \Phi(I)$ using the same protocol as for parts.

\paragraph{Results.}

\Cref{tab:eval-part-composition} shows that our method achieves performance comparable to directly reconstructing the objects using the RM ($\hat \object = \Phi(I)$), with the additional benefit of producing the reconstruction structured into parts, which are useful for downstream applications such as editing.

\subsection{Applications}%
\label{sec:applications}

\paragraph{Part-aware text-to-3D generation.}

First, we apply \method to part-aware text-to-3D generation.
We train a text-to-multi-view generator similar to~\cite{siddiqui24meta}, which takes a text prompt as input and outputs a grid of four views.
For illustration, we use the prompts from DreamFusion~\cite{poole23dreamfusion:}.
As shown in \cref{fig:application}, \method can effectively generate 3D objects with distinct and completed parts, even in challenging cases with heavy occlusions, such as the gummy bear. 
Additional examples are provided in the supp.~mat.

\paragraph{Part-aware image-to-3D generation.}

Next, we consider part-aware image-to-3D generation using images from \cite{xu24instantmesh:,han2025vfusion3d}.
Building upon the text-to-multi-view generator, we further fine-tune the generator to accept images as input with a strategy similar to~\cite{ye2023ip-adapter}. Further training details are provided in supp.~mat.
Results are shown in \cref{fig:application} demonstrating that \method is successful in this case as well.

\paragraph{Real-world 3D object decomposition.}

\method can also decompose real-world 3D objects.
We show this using objects from Google Scanned Objects (GSO)~\cite{downs2022google} for this purpose.
Given a 3D object from GSO, we render different views to obtain a an image grid and then apply \method as above.
The last row of \Cref{fig:application} shows that \method can effectively decompose real-world 3D objects too.

\paragraph{3D part editing.}

Finally, we show that once the 3D parts are decomposed, they can be further modified through text input.
As illustrated in \cref{fig:3d_editing}, a variant of our method enables effective editing of the shape and texture of the parts based on textual prompts.
The details of the 3D editing model are provided in supplementary materials.

%% file: tables/tab_eval_part_segmentation.tex
\begin{table}[t]
\centering
\setlength{\tabcolsep}{4pt}
\resizebox{\linewidth}{!}{
\begin{tabular}{lcccc}
\toprule

& \multicolumn{2}{c}{\textbf{Automatic}}  & \multicolumn{2}{c}{\textbf{Seeded}} \\
Method & $\text{mAP}_{50}$$\uparrow$ & $\text{mAP}_{75}$$\uparrow$ & $\text{mAP}_{50}$$\uparrow$ & $\text{mAP}_{75}$$\uparrow$ \\ \midrule 
Part123~\cite{liu24part123:}                     & 11.5  & 7.4  & 10.3   & 6.5  \\
\textbf{}$\text{SAM2}^{\dag}$~\cite{ravi24sam-2}          & 20.3  & 11.8  & 24.6   & 13.1  \\
$\text{SAM2}^*$~\cite{ravi24sam-2}               & 37.4  & 27.0  & 44.2   & 30.1  \\
$\text{SAM2}$~\cite{ravi24sam-2}                 & 35.3  & 23.4  & 41.4   & 27.4  \\
\rowcolor{blue!10} $\text{\method (1 sample)}$   & 45.2 & 32.9 & 44.9  & 33.5 \\
\rowcolor{blue!10} $\text{\method (5 samples)}$  & 54.2 & 33.9 &  51.3 & 32.9 \\
\rowcolor{blue!10} $\text{\method (10 samples)}$ & \textbf{59.3} & \textbf{38.5} & \textbf{53.7}  & \textbf{35.4} \\
\bottomrule
\end{tabular}
}
\vspace{-2mm}
\caption{\textbf{Segmentation results.}
$\text{SAM2}^*$ is fine-tuned our data and $\text{SAM2}^{\dag}$ is fine-tuned for multi-view segmentation.}%
\label{tab:eval-part-segmentation}
\vspace{-3mm}
\end{table}

%% file: tables/tab_eval_part_completion.tex
\begin{table*}[tb]
\newcommand{\bul}{{\tiny\ding{108}}}
\centering

\setlength{\tabcolsep}{4pt}
\resizebox{0.85\textwidth}{!}{
\begin{tabular}{lccccccccc}
\toprule
&&&&
\multicolumn{3}{c}{\textbf{View completion $J$}} &
\multicolumn{3}{c}{\textbf{3D reconstruction $\objectpart$}}
\\
\textbf{Method} & \textbf{Compl.} & \textbf{Multi-view} & \textbf{Context} &
CLIP$\uparrow$ & LPIPS$\downarrow$ & PSNR$\uparrow$ & CLIP$\uparrow$ & LPIPS$\downarrow$ & PSNR$\uparrow$   \\ \midrule
Oracle ($\hat J = J$) & GT & --- & --- & 1.0   & 0.0   & $\infty$ & 0.957 & 0.027 & 18.91  \\
\midrule
\rowcolor{blue!10} \method ($\hat J = \mathcal{B}(I \odot M,I)$) & \ding{51} & \ding{51} & \ding{51} & \textbf{0.974} & \textbf{0.015} & \textbf{21.38}    & \textbf{0.936} & \textbf{0.039} &\textbf{17.16}  \\
\enspace w/o context$^\dag$  ($\hat J = \mathcal{B}(I \odot M)$) & \ding{51} & \ding{51} & \ding{55} & 0.951 & 0.028 & 16.80    & 0.923 & 0.046 & 14.83 \\
\enspace single view$^\ddag$ ($\hat J_v = \mathcal{B}(I_v \odot M_v,I_v)$) & \ding{51} & \ding{55} & \ding{51} & 0.944 & 0.031 & 15.92    & 0.922 & 0.051 & 13.25  \\ \midrule

None ($\hat J = I \odot M$) & \ding{55} & --- & --- & 0.932 & 0.039 & 13.24 & 0.913 & 0.059 & 12.32  \\ 
\bottomrule
\end{tabular}
}
\vspace{-1mm}
\caption{\textbf{{Part completion results}.}
We first evaluate view part completion by computing scores w.r.t. the ground-truth multi-view part image $J$. Then, we evaluate 3D part reconstruction by reconstructing each part $\objectpart$ and rendering it. See text for details.}
\label{tab:eval-part-completion}
\vspace{-2mm}

\end{table*}

%% file: figures/fig_part_completion.tex
\begin{figure}[t]
  \centering
  \includegraphics[width=\linewidth]{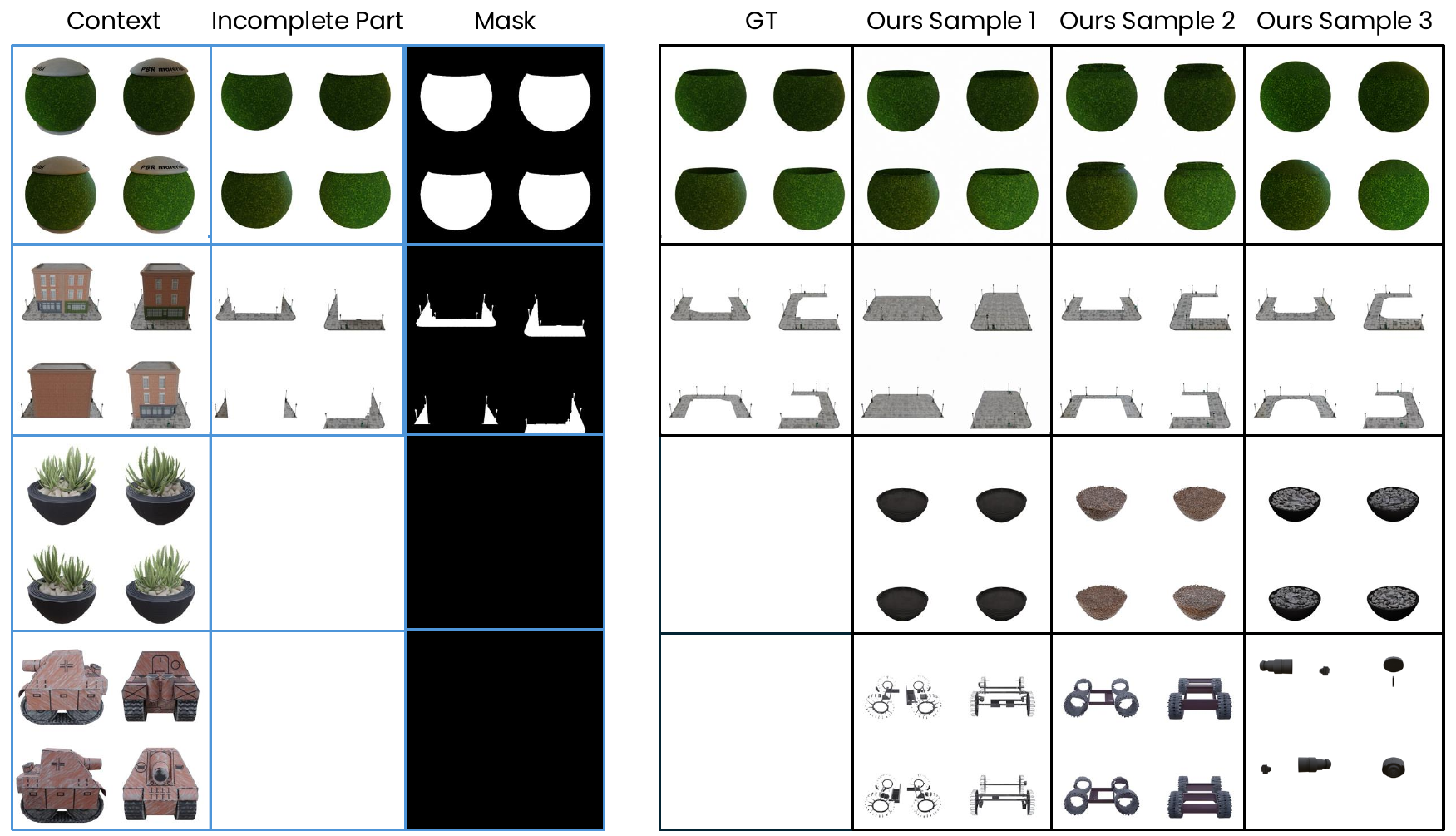}
  \caption{\textbf{Qualitative results of part completion.} The images with blue borders are the inputs. Our algorithm produces various plausible outputs across different runs. Even if given an empty part, \method attempts to generate internal structures inside the object, such as sand or inner wheels. }%
  \label{fig:part_completion}
  \vspace{-4mm}
\end{figure}

%% file: figures/fig_application.tex
\begin{figure*}[t]
\centering
\includegraphics[width=0.99\linewidth]{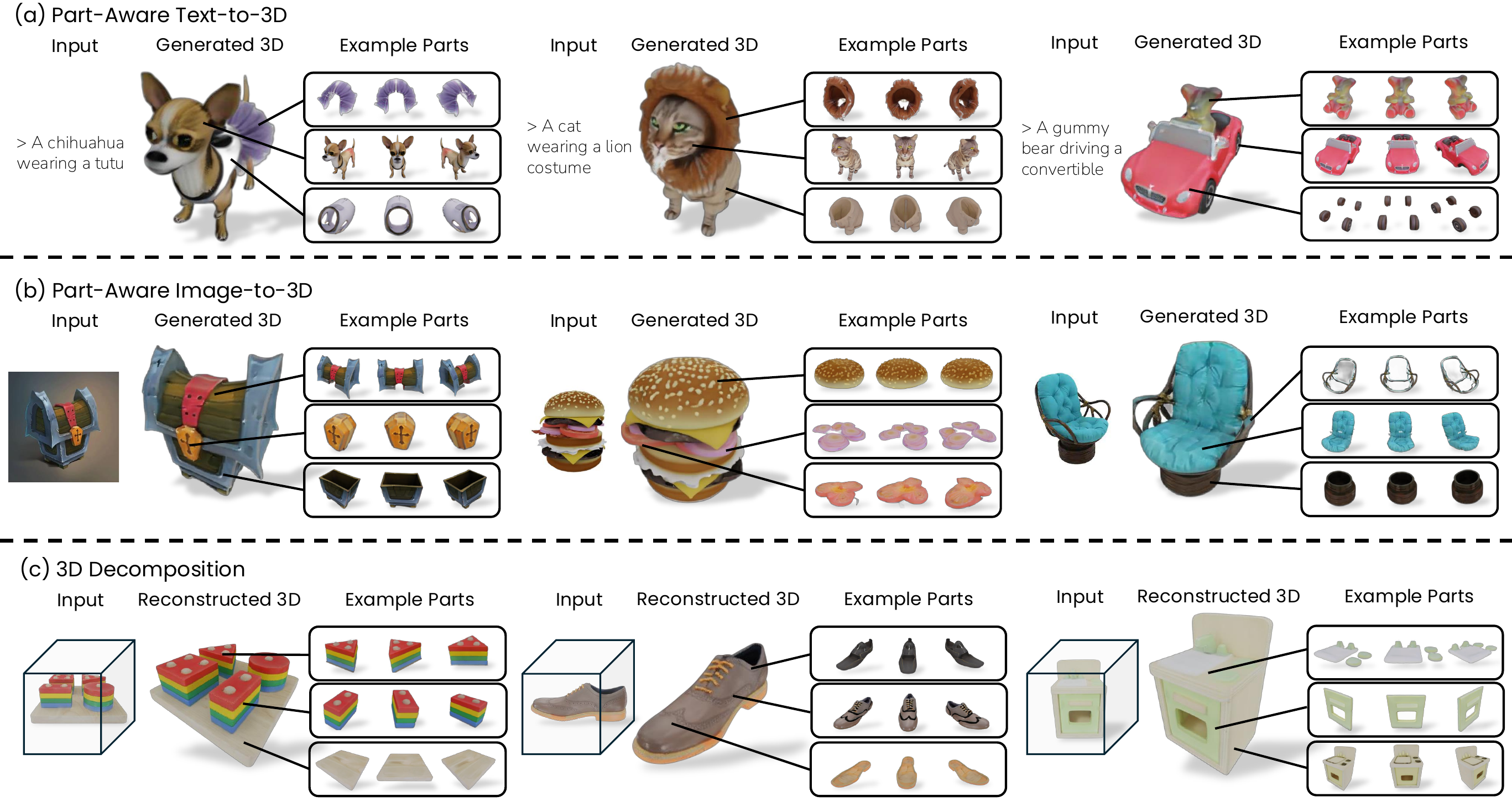}
\vspace{-.8em}
\caption{\textbf{Examples of applications.} \method can effectively generate or reconstruct 3D objects with meaningful and realistic parts in different scenarios: a) Part-aware text-to-3D generation; b) Part-aware image-to-3D generation; c) 3D decomposition.}%
\label{fig:application}
\vspace{-3mm}
\end{figure*}

%% file: figures/fig_3d_editing.tex
\begin{figure}[t]
  \centering
  \includegraphics[width=0.89\linewidth]{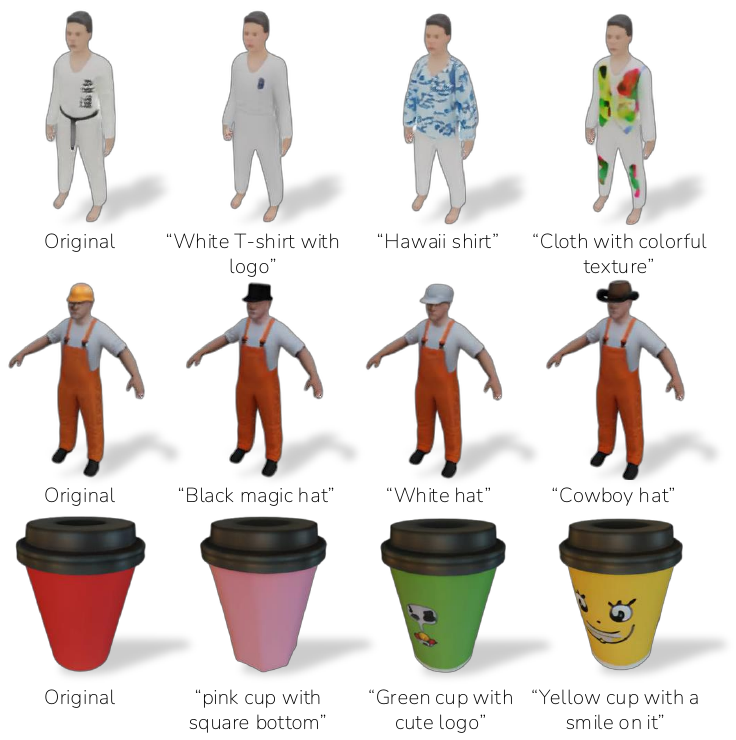}
  \vspace{-.8em}
  \caption{\textbf{3D part editing.} We can edit the appearance and shape of the 3D objects with text prompt.}%
  \label{fig:3d_editing}
\end{figure}

%% file: tables/tab_eval_part_composition.tex
\begin{table}[t]
\centering
\setlength{\tabcolsep}{2.7pt}
\begin{tabular}{lccc}
\toprule
Method                                                                & CLIP$\uparrow$ & LPIPS$\downarrow$ & PSNR$\uparrow$ \\ \midrule
\rowcolor{blue!10} \method ($\hat\object = \bigcup_k \Phi(\hat J_k)$) & 0.952                & 0.065                & 20.33             \\
Unstructured ($\hat\object = \Phi(I)$)                                & 0.955                & 0.064              & 20.47           \\ 
\bottomrule
\end{tabular}
\vspace{-1mm}
\caption{\textbf{Model reassembling result.}
The quality of 3D reconstruction of the object as a whole is close to that of the part-based compositional reconstruction, which proves that the predicted parts fit together well.}%
\vspace{-3mm}
\label{tab:eval-part-composition}
\end{table}

%% file: sec/5_conclusion.tex
\section{Conclusion}%
\label{sec:conclusion}

We have introduced \method, a novel approach to generate or reconstruct compositional 3D objects from text, images, or unstructured 3D objects.
\method can reconstruct in 3D parts that are even minimally visible, or not visible at all, utilizing the guidance of a specially-designed multi-view diffusion prior.
We have also shown several application of \method, including text-guided part editing.
This is a promising step towards the generation of 3D assets that are more useful in professional workflows.

%% file: sec/X_suppl.tex
\clearpage
\setcounter{page}{1}
\maketitlesupplementary
\appendix

This supplementary material contains the following parts:
\begin{itemize}
    \item \textbf{Implementation Details.} Detailed descriptions of the training and inference settings for all models used in \method are provided.

    \item \textbf{Additional Experiment Details.} We describe the detailed evaluation metrics employed in the experiments and provide additional experiments.

    \item \textbf{Additional Examples.} We include more outputs of our method, showcasing applications with part-aware text-to-3D, part-aware image-to-3D, real-world 3D decomposition, and iteratively adding parts.

    \item \textbf{Failure Case.} We analyse the modes of of failure of \method. 

    \item \textbf{Ethics and Limitation.} We provide a discussion on the ethical considerations of data and usage, as well as the limitations of our method.

\end{itemize}

\section{Implementation Details}

We provide the details of training used in \method (\Cref{ssec:t2mv,ssec:i2mv,ssec:mvseg,ssec:mvcomplete}). In addition, we provide the implementation details for the applications: for part composition (\Cref{ssec:assembly}) and for part editing (\Cref{ssec:edit}).

\subsection{Text-to-multi-view generator}
\label{ssec:t2mv}
We fine-tune the text-to-multi-view generator starting with a pre-trained text-to-image diffusion model trained on billions of image-text pairs that uses an architecture and data similar to Emu~\cite{dai23emu:}. We change the target image to a grid of $2\times2$ views as described in Section 3.5 following Instant 3D \cite{li24instant3d:} via v-prediction~\cite{salimans2022progressive} loss. The resolution of each view is $512\times512$, resulting in the total size of $1024\times1024$. To avoid the problem of the cluttered background mentioned in~\cite{li24instant3d:}, we rescale the noise scheduler to force a zero terminal signal-to-noise ratio (SNR) following~\cite{lin2024common}. We use the DDPM scheduler with 1000 steps~\cite{ho20denoising} for training. During the inference, we use DDIM~\cite{song21denoising} scheduler with 250 steps. The model is trained with 64 H100 GPUs with a total batch size of 512 and a learning rate $10^{-5}$ for 10k steps.

\begin{figure}
    \centering
    \includegraphics[width=\linewidth]{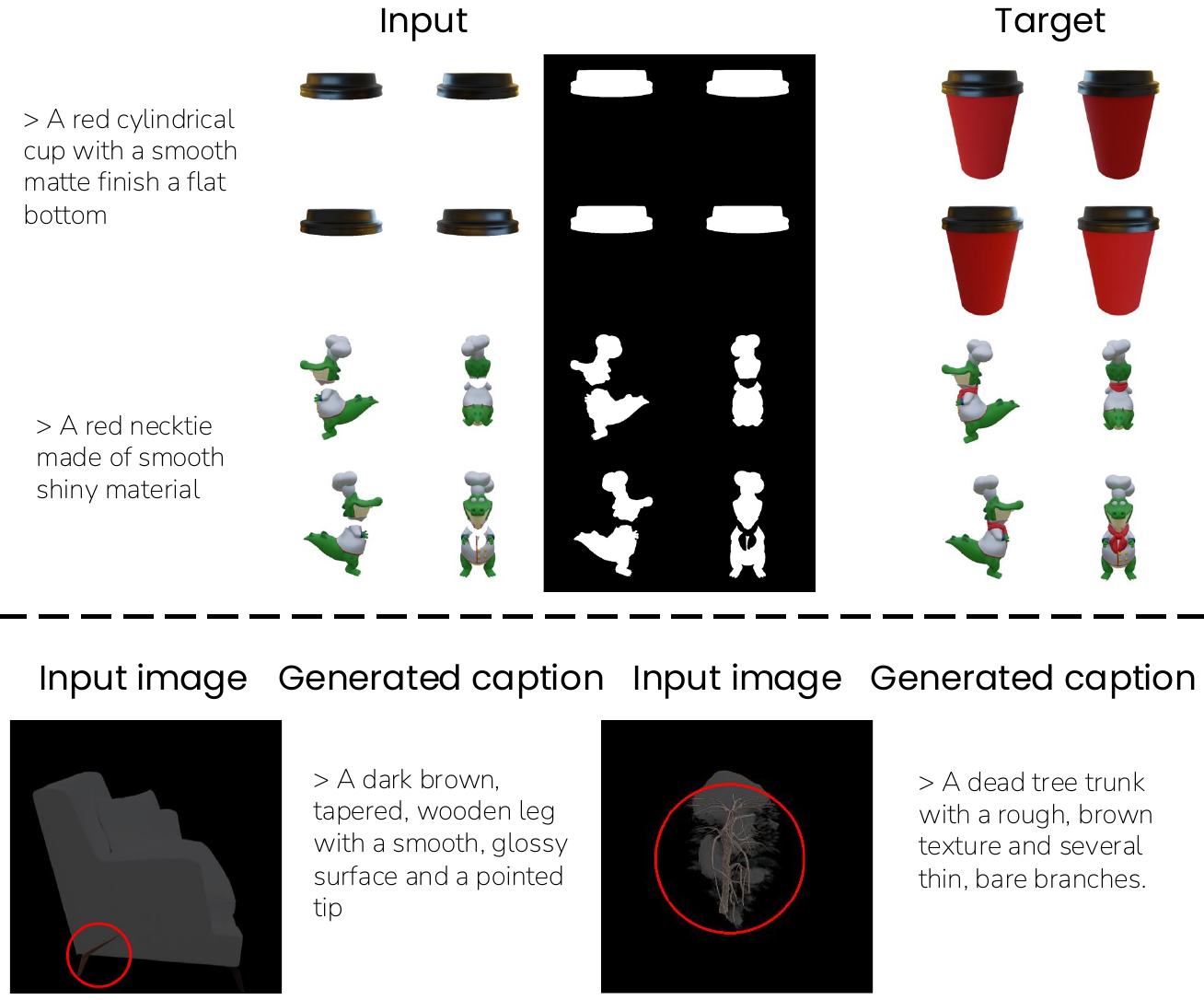}
    \caption{\textbf{3D part editing and captioning examples.} The top section illustrates training examples for the editing network, where a mask, a masked image, and text instructions are provided as conditioning to the diffusion network, which fills in the part based on the given textual input. The bottom section demonstrates the input for the part captioning pipeline. Here, a red circle and highlights are used to help the large vision-language model (LVLM) identify and annotate the specific part.}%
    \label{fig:3d_editing_supp}
\end{figure}
\begin{figure}[ht]
    \centering
    \begin{subfigure}[b]{0.49\textwidth}
        \centering
        \includegraphics[width=\textwidth]{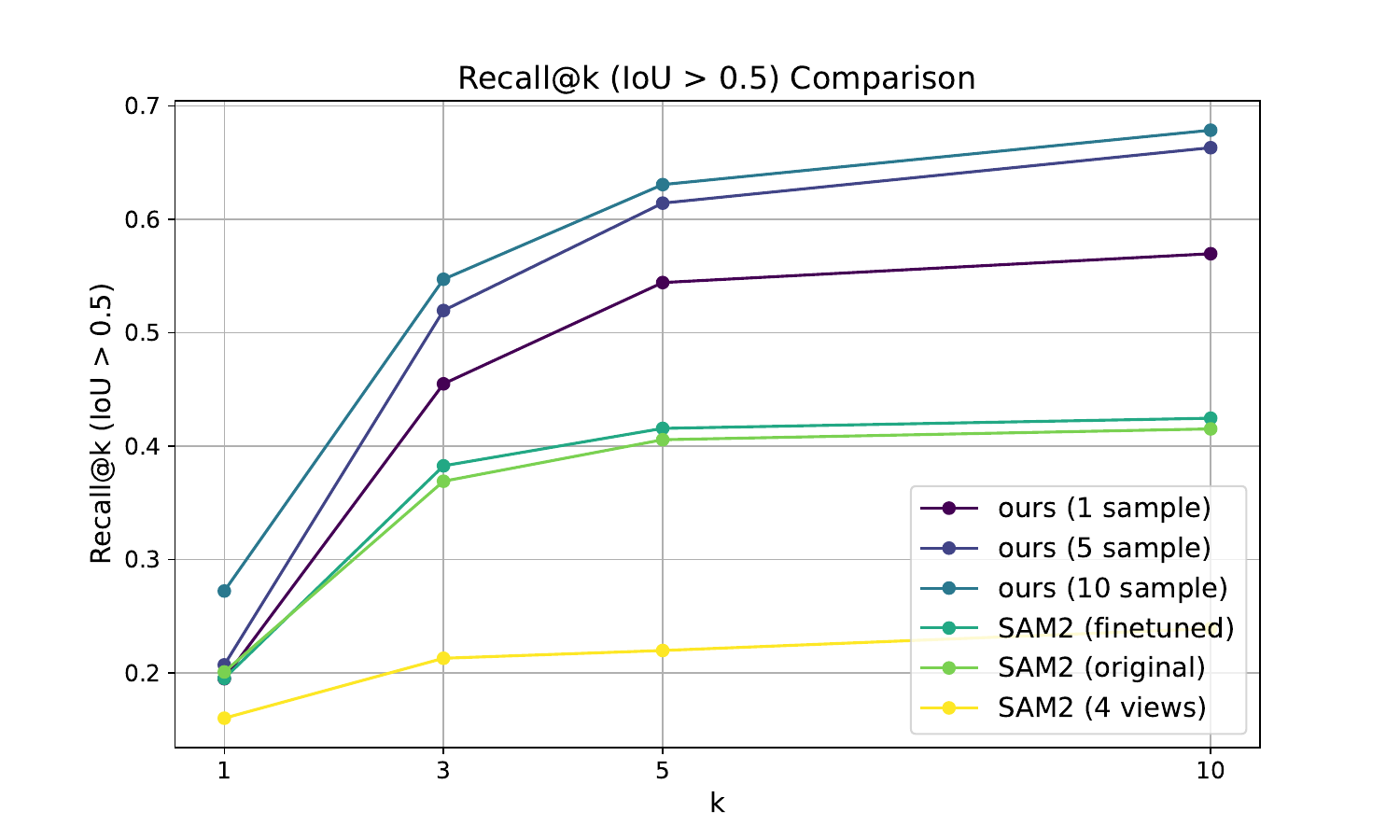} %
    \end{subfigure}
    \begin{subfigure}[b]{0.49\textwidth}
        \centering
        \includegraphics[width=\textwidth]{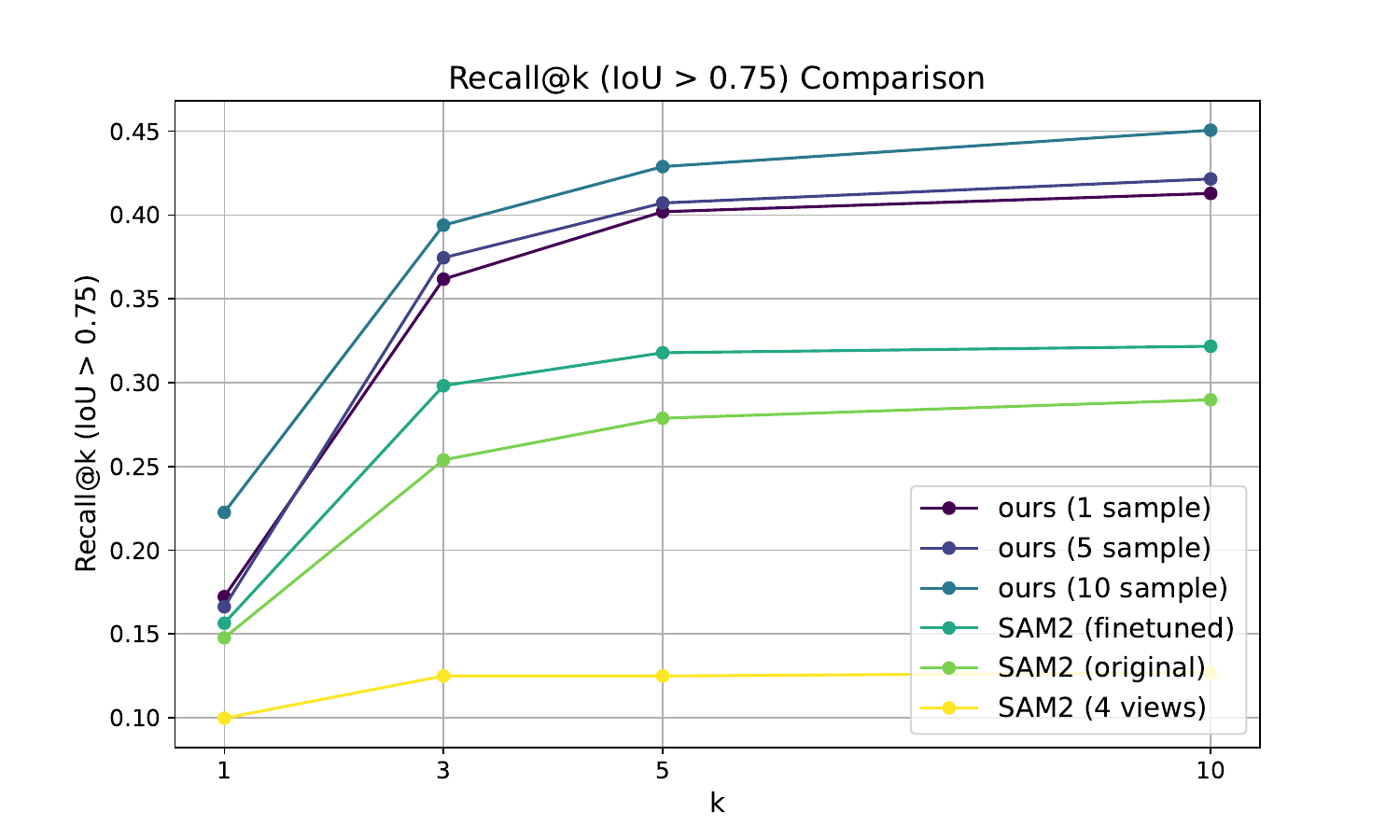} %
    \end{subfigure}
    
    \caption{\textbf{Recall curve of different methods.} Our method achieve better performance comparing with SAM2 and its variants.}
    \label{fig:recall_curve}
\end{figure}
\begin{figure*}
    \centering
    \includegraphics[width=\linewidth]{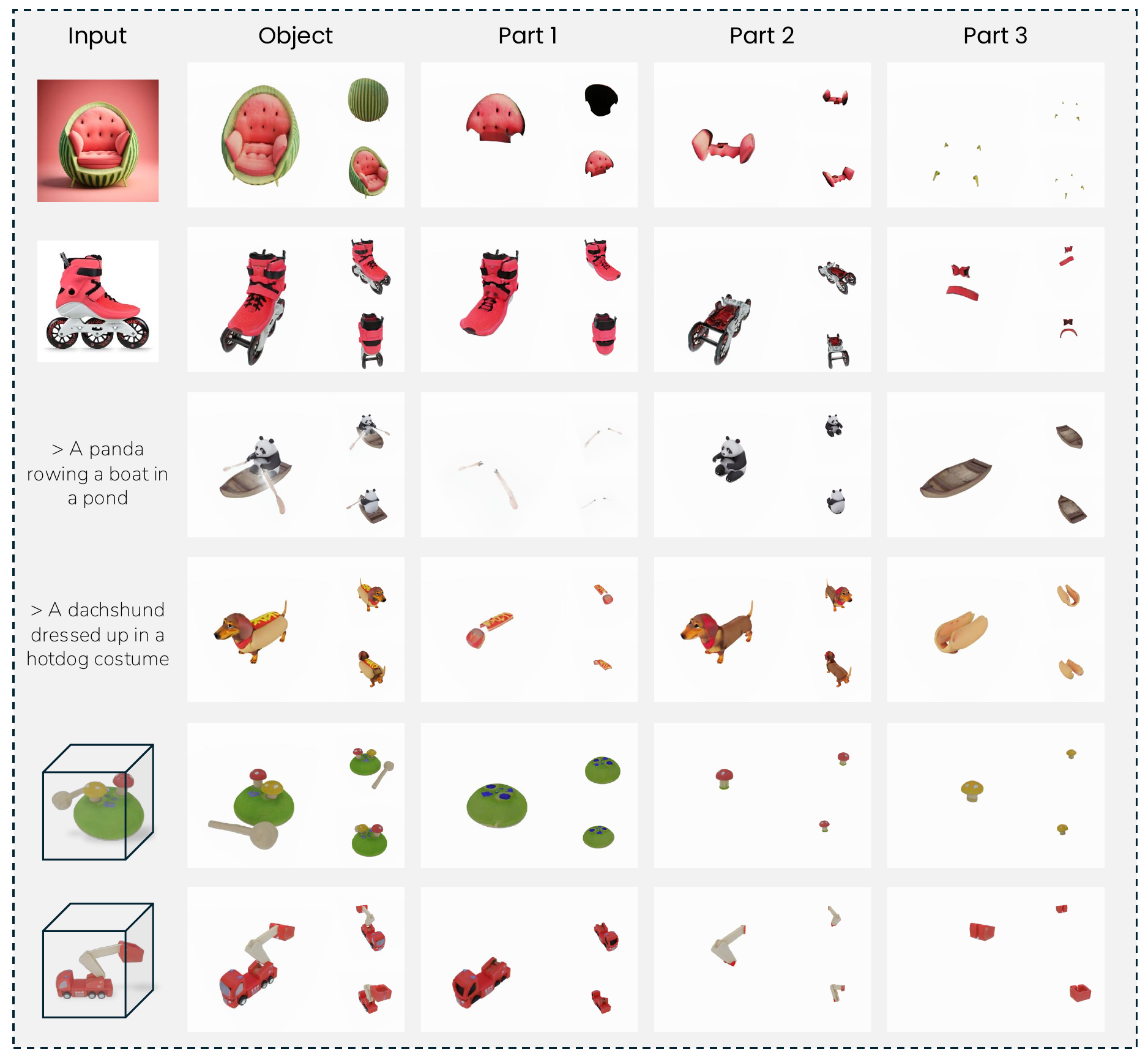}
    \caption{\textbf{More examples.} Additional examples illustrate that \method can process various modalities and effectively generate or reconstruct 3D objects with distinct parts.}%
    \label{fig:more_examples}
\end{figure*}

\begin{figure*}
    \centering
    \includegraphics[width=\linewidth]{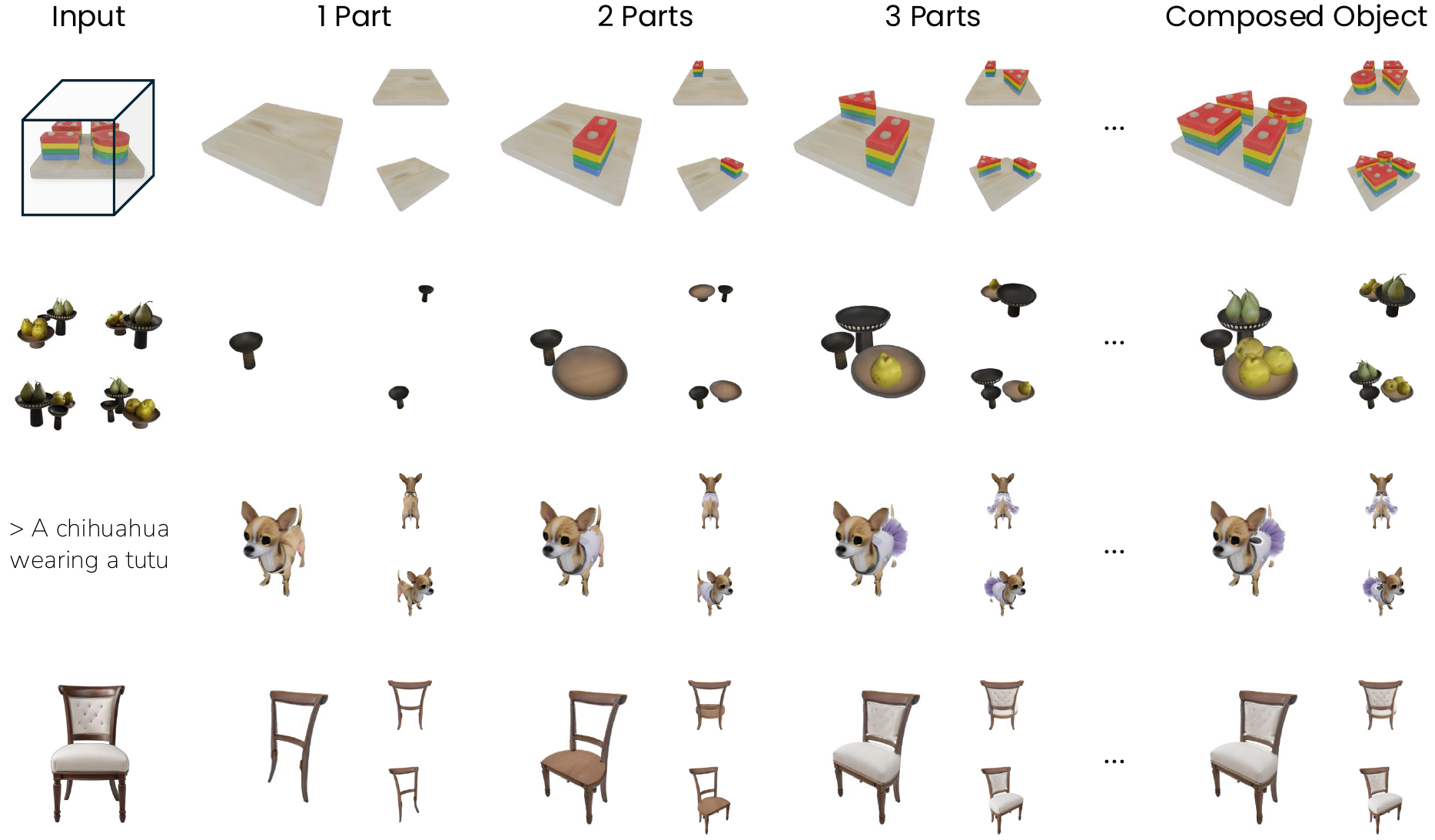}
    \caption{\textbf{Iteratively adding parts.} We show that users can iteratively add parts and combine the results of \method pipeline.}%
    \label{fig:iterative_add_parts}
\end{figure*}
\begin{figure}
    \centering
    \includegraphics[width=\linewidth]{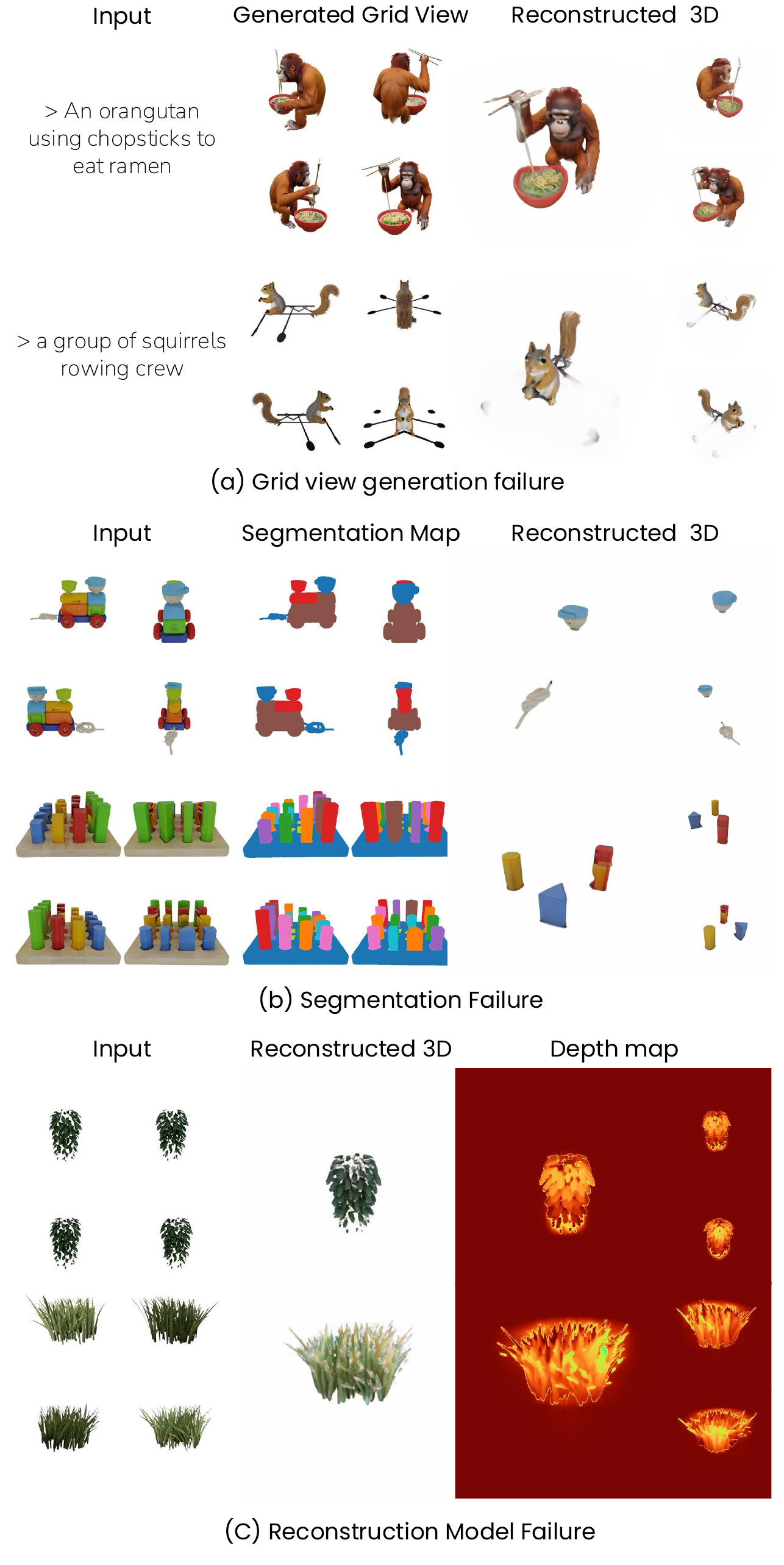}
    \caption{\textbf{Failure Cases.} (a) Multi-view grid generation failure, where the generated views lack 3D consistency. (b) Segmentation failure, where semantically distinct parts are incorrectly grouped together. (c) Reconstruction model failure, where the complex geometry of the input leads to inaccuracies in the depth map.}%
    \label{fig:failure_case}
\end{figure}

\subsection{Image-to-multi-view generator}
\label{ssec:i2mv}
Building on the text-to-multi-view generator, we further fine-tune the model to accept images as input conditioning instead of text. The text condition is removed by setting it to a default null condition (an empty string). We concatenate the conditional image to the noised image along the spatial dimension, following~\cite{chong2024catvton}. Additionally, inspired by IP-adapter~\cite{ye2023ip-adapter}, we introduce another cross-attention layer into the diffusion model. The input image is first converted into tokens using CLIP \cite{radford21learning}, then reprojected into 157 tokens of dimension 1024 using a Perceiver-like architecture~\cite{jaegle22perceiver}. To train the model, we utilize all 140k 3D models of our data collection, selecting conditional images with random elevation and azimuth but fixed camera distance and field of view. We use the DDPM scheduler with 1000 steps~\cite{ho20denoising}, rescaled SNR, and v-prediction for training. Training is conducted with 64 H100 GPUs, a batch size of 512, and a learning rate of $10^{-5}$ over 15k steps. 

\subsection{Multi-view segmentation network}
\label{ssec:mvseg}
To obtain the multi-view segmentation network, we also fine-tune the pre-trained text-to-multi-view model. The input channels are expanded from 8 to 16 to accommodate the additional image input, where 8 corresponds to the latent dimension of the VAE used in our network. We create segmentation-image pairs as inputs. The training setup follows a similar recipe to that of the image-to-multi-view generator, employing a DDPM scheduler, v-prediction, and rescaled SNR. The network is trained with 64 H100 GPUs, a batch size of 512, a learning rate of $10^{-5}$, for 10k steps.

\subsection{Multi-view completion network}
\label{ssec:mvcomplete}
The training strategy for the multi-view completion network mirrors that of the multi-view segmentation network, with the key difference in the input configuration. The number of input channels (in latent space) is increased to 25 by including the context image, masked image, and binary mask, where the mask remains a single unencoded channel. Example inputs are illustrated in Figure 5 of the main text. The network is trained with 64 H100 GPUs, a batch size of 512, a learning rate of $10^{-5}$, and for approximately 10k steps.

\subsection{Parts assembly}
\label{ssec:assembly}
When compositing an object from its parts, we observed that simply combining the implicit neural fields of parts reconstructed by the Reconstruction Model (RM) in the rendering process with their respective spatial locations achieves satisfactory results.

To describe this formally, we first review the rendering function of LightplaneLRM \cite{cao2024lightplane} that we use as our reconstruction model. LightplaneLRM employs a generalized Emission-Absorption (EA) model for rendering, which calculates transmittance $T_{ij}$, representing the probability of a photon emitted at position $x_{ij}$ (the $j_{th}$ sampling point in the $i_{th}$ ray) reaching the sensor. Then the rendered feature (\eg color) $v_{i}$ of ray $r_{i}$ is computed as:
$$ v_i = \sum_{j=1}^{R-1}(T_{i,j-1} - T_{i,j})f_{v}(x_{ij})$$
where $f_{v}(x_{ij})$ denotes the feature of the 3D point $x_{ij}$; $T_{i,j} = \text{exp}(-\sum_{k=0}^{j}\Delta \cdot \sigma(x_{ik}))$, where $\Delta$ is the distance between two sampled points and $\sigma(x_{ik})$ is the opacity at position $x_{ik}$, $T_{i,j-1} - T_{i,j}$ captures the visibility of the point.

Now we show how we generalise it to rendering $N$ parts. Given feature functions $f_{v}^{1}, \dots, f_{v}^{N}$ and their opacity functions $\sigma^1, \cdots, \sigma^N$, the rendered feature of a specific ray $r_i$ becomes:
$$ v_i = \sum_{j=1}^{R-1}\sum_{h=1}^{N}(\hat T_{i,j-1} - \hat T_{i,j})w_{ij}^h \cdot f_{v}^{h}(x_{ij}).$$
where $w_{ij}^h = \sigma^h(x_{ij})/\sum_{l=1}^N{\sigma^l(x_{ij})}$ is the weight of the feature $f_{v}^{h}(x_{ij})$ at $x_{ij}$ for part $h$; $\hat T_{i,j} = \text{exp}(-\sum_{k=0}^{j}\sum_{h=1}^{N}\Delta \cdot \sigma^{h}(x_{ik}))$, $\Delta$ is the distance between two sampled points and $\sigma^h(x_{ik})$ is the opacity at position $x_{ik}$ for part $h$, and $\hat T_{i,j-1} - \hat T_{i,j}$ is the visibility of the point.  

\subsection{3D part editing}
\label{ssec:edit}
As shown in the main text and Figure 7, once 3D assets are generated or reconstructed as a composition of different parts through \method, specific parts can be edited using text instructions to achieve 3D part editing. To enable this, we fine-tune the text-to-multi-view generator using part multi-view images, masks, and text description pairs. Example of the training data are shown in \Cref{fig:3d_editing_supp} (top). Notably, instead of supplying the mask for the part to be edited, we provide the mask of the remaining parts. This design choice encourages the editing network to imagine the part’s shape without constraining the region where it has to project. The training recipe is similar to multi-view segmentation network.

To generate captions for different parts, we establish an annotation pipeline similar to the one used for captioning the whole object, where captions for various views are first generated using LLAMA3 and then summarized into a single unified caption using LLAMA3 as well. The key challenge in this variant is that some parts are difficult to identify without knowing the context information of the object. We thus employ the technique inspired by~\cite{shtedritski2023does}. Specifically, we use red annulet and alpha blending to emphasize the part being annotated. Example inputs and generated captions are shown in \Cref{fig:3d_editing_supp} (bottom). The network is trained with 64 H100 GPUs, a batch size of 512, and the learning rate of $10^{-5}$ over 10,000 steps.

\section{Additional Experiment Details}

We provide a detailed explanation of the ranking rules applied to different methods and the formal definition of mean average precision (mAP) used in our evaluation protocol. Additionally, we report the recall at $K$ in the automatic segmentation setting.

\paragraph{Ranking the parts.}

For evaluation using mAP and recall at $K$, it is necessary to rank the part proposal.
For our method, we run the segmentation network several times and concatenate the results into an initial set
$\mathcal{P}$ of segment proposals.
Then, we assign to each segment $\hat M \in \mathcal{P}$ a reliability score based on how frequently it  overlaps with similar segments in the list, \ie,
$$
s(\hat M) =
\left | 
\left \{
\hat M' \in \mathcal{P}: 
m
(\hat M ', \hat M) > \frac{1}{2}
\right \}
\right|
$$
where the \emph{Intersection over Union} (IoU)~\cite{larlus06pascal} metric is given by:
$$
m(\hat M, M)
= \operatorname{IoU}(\hat M, M)
= 
\frac
{|\hat M \cap M|+\epsilon}
{|\hat M \cup M|+\epsilon}.
$$
The constant $\epsilon=10^{-4}$ smooths the metric when both regions are empty, in which case $m(\phi, \phi) = 1$, and will be useful later.

Finally, we sort the regions $M$ by decreasing score $s(M)$ and, scanning the list from high to low, we incrementally remove duplicates down the list if they overlap by more than $1/2$ with the regions selected so far.
The final result is a ranked list of multi-view masks
$
\mathcal{M}=(\hat M_{1},\ldots,\hat M_{N})
$
where $N \le |\mathcal{P}|$ and:
$$
\forall i < j:
~~
s(\hat M_i) \ge s(\hat M_j)
~\wedge~
m(\hat M_i, \hat M_j) < \frac{1}{2}.
$$

Other algorithms like SAM2 come with their own region reliability metric $s$, which we use for sorting.
We otherwise apply non-maxima suppression to their ranked regions in the same way as ours.

\paragraph{Computing mAP.}

The image $I$ comes from an object $\object$ with parts 
$
(\objectpart^1,\dots,\objectpart^S)
$
from which we obtain the ground-truth part masks
$
\mathcal{S}
=
(M^1,\dots,M^S)
$
as explained in Section 3.5 in the main text.
We assign ground-truth segments to candidates following the procedure: we go through the list
$
\mathcal{M} = (\hat M_{1},\ldots,\hat M_{N})
$
and match the candidates one by one to the ground truth segment with the highest IOU, exclude that ground-truth segment, and continue traversing the candidate list.
We measure the degree of overlap between a predicted segment and a ground truth segment as $m(\hat M, M)\in[0,1]$.
Given this metric, we then report the \emph{mean Average Precision} (mAP) metric at different IoU thresholds $\tau$.
Recall that, based on this definition, computing the AP curve for a sample involves matching predicted segments to ground truth segments in ranking order, ensuring that each ground truth segment is matched only once, and considering any unmatched ground truth segments.

In more detail, we start by scanning the list of segments $\hat M_k$ in order $k=1,2,\dots$.
Each time, we compare $\hat M_k$ to the ground truth segments $\mathcal{S}$ and define:
$$
s^* = \operatornamewithlimits{argmax}_{s=1,\dots,S} m(\hat M_k, M_s).
$$
If
$
m(\hat M_k, M_{s^*}) \geq \tau,
$
then we label the region $M_s$ as retrieved by setting $y_k=1$ and removing $M_s$ from the list of ground truth segments not yet recalled by setting
$$
\mathcal{S}\leftarrow \mathcal{S} \setminus \{M_{s^*}\}.
$$
Otherwise, 
if
$
m(\hat M_k, M_{s^*}) < \tau
$
or if $\mathcal{S}$ is empty, we set $y_k=0$.
We repeat this process for all $k$, which results in labels
$(y_1,\dots,y_N) \in \{0,1\}^{N}$.
We then set the \emph{average precision} (AP) at $\tau$ to be:
$$
\operatorname{AP}(\mathcal{M}, \mathcal{S};\tau)
=
\frac{1}{S}
\sum_{k=1}^N
\sum_{i=1}^k
\frac{y_i y_k}{k}.
$$
Note that this quantity is at most $1$ because by construction $\sum_{i=1}^N y_i \leq S$  as we cannot match more proposal than there are ground truth regions.
mAP is defined as the average of the AP over all test samples.

\paragraph{Computing recall at $K$.}

For a given sample, we define \emph{recall at $K$} the curve
$$
R(K;\mathcal{M}, \mathcal{S}, \tau) =
\frac{1}{S}
\sum_{s=1}^S
\chi\left(
  \max_{k=1,\dots,K} m(\hat M_s, M_k) > \tau
\right).
$$
Hence, this is simply the fraction of ground truth segments recovered by looking up to position $K$ in the ranked list of predicted segments. The results in \Cref{fig:recall_curve} demonstrate that our diffusion-based method outperforms SAM2 and its variants by a large margin and shows consistent improvement as the number of samples increases.

\paragraph{Seeded part segmentation.}

To evaluate \emph{seeded part segmentation}, the assessment proceeds as before, except that a single ground truth part $\objectpart$ and mask $M$ is considered at a time, and the corresponding seed point $u \in M$ is passed to the algorithm $(\hat M_1,\dots,\hat M_K)=\mathcal{A}(I,u)$.
Note that, because the problem is still ambiguous, it makes sense for the algorithm to still produce a ranked list of possible part segments.

\section{Additional Examples}%
\label{sed:additional_examples} 

\paragraph{More application examples.} We provide additional application examples in \Cref{fig:more_examples}, showcasing the versatility of our approach to varying input types. These include part-aware text-to-3D generation, where textual prompts guide the synthesis of 3D models with semantically distinct parts; part-aware image-to-3D generation, which reconstructs 3D objects from a single image while maintaining detailed part-level decomposition; and real-world 3D decomposition, where complex real-world objects are segmented into different parts. These examples demonstrate the broad applicability and robustness of \method in handling diverse inputs and scenarios.

\paragraph{Iteratively adding parts.} As shown in \Cref{fig:iterative_add_parts}, we demonstrate the capability of our approach to compose a 3D object by iteratively adding individual parts to it. Starting with different inputs, users can seamlessly integrate additional parts step by step, maintaining consistency and coherence in the resulting 3D model. This process highlights the flexibility and modularity of our method, enabling fine-grained control over the composition of complex objects while preserving the semantic and structural integrity of the composition.

\section{Failure Cases}
As outlined in the method section, \method incorporates several steps, including multi-view grid generation, multi-view segmentation, multi-view part completion, and 3D part reconstruction. Failures at different stages will result in specific issues. For instance, as shown in \Cref{fig:failure_case}(a), failures in grid view generation can cause inconsistencies in 3D reconstruction, such as misrepresentations of the orangutan's hands or the squirrel's oars. The segmentation method can sometimes group distinct parts together, and limited, in our implementation, to objects containing no more than 10 parts, otherwise it merges different building blocks into a single part. Furthermore, highly complex input structures, such as dense grass and leaves, can lead to poor reconstruction outcomes, particularly in terms of depth quality, as illustrated in \Cref{fig:failure_case}(c).

\section{Ethics and Limitation}
\paragraph{Ethics.}

Our models are trained on datasets derived from artist-created 3D assets. These datasets may contain biases that could propagate into the outputs, potentially resulting in culturally insensitive or inappropriate content.
To mitigate this, we strongly encourage users to implement safeguards and adhere to ethical guidelines when deploying \method in real-world applications.

\paragraph{Limitation.} In this work, we focus primarily on object-level generation, leveraging artist-created 3D assets as our training dataset. However, this approach is heavily dependent on the quality and diversity of the dataset. Extending the method to scene-level generation and reconstruction is a promising direction but it will require further research and exploration.